\documentclass{article}

\usepackage{cite}
\usepackage[margin=3cm]{geometry}
\usepackage{graphicx}
\usepackage{hyperref}
\usepackage{float}
\usepackage{chngcntr}
\usepackage[table]{xcolor}
\usepackage{colortbl}

\usepackage{setspace}

\begin{document}
\doublespacing
\title{
% Enhancing Gait Recognition Accuracy using Body Landmarks and Bidirectional Siamese Recurrent Neural Network
A Bidirectional Siamese Recurrent Neural Network for Accurate Gait Recognition Using Body Landmarks
%Siamese BiGRU-dualStack: An Accurate Network for Enhancing Gait Recognition Accuracy using Body Landmarks
}

\author{Proma Hossain Progga$^1$, Md. Jobayer Rahman$^1$, Swapnil Biswas$^1$, Md. Shakil Ahmed$^1$,\\ Arif Reza Anwary$^2$ and Swakkhar Shatabda$^3$
%\thanks{Department of Computer Science Engineering, United International University, Dhaka 1212, Bangladesh}
\\
$^1$Department of Computer Science Engineering, United International University, \\
Dhaka 1212, Bangladesh\\
$^2$School of Computing, Edinburgh Napier University, United Kingdom\\
$^3$Department of Computer Science Engineering, BRAC  University, \\
Dhaka 1212, Bangladesh\\}
\date{}

% \markboth{IEEE Sensors}%
% {Shell \MakeLowercase{\textit{et al.}}: Bare Demo of IEEEtran.cls for IEEE Journals}

\maketitle

\begin{abstract}
Gait recognition is a significant biometric technique for person identification, particularly in scenarios where other physiological biometrics are impractical or ineffective. In this paper, we address the challenges associated with gait recognition and present a novel approach to improve its accuracy and reliability. The proposed method leverages advanced techniques, including sequential gait landmarks obtained through the Mediapipe pose estimation model, Procrustes analysis for alignment, and a Siamese biGRU-dualStack Neural Network architecture for capturing temporal dependencies. Extensive experiments were conducted on large-scale cross-view datasets to demonstrate the effectiveness of the approach, achieving high recognition accuracy compared to other models. The model demonstrated accuracies of 95.7$\%$, 94.44$\%$, 87.71$\%$, and \textcolor{black}{86.6$\%$} on CASIA-B, SZU RGB-D, OU-MVLP, and \textcolor{black}{Gait3D }datasets respectively. The results highlight the potential applications of the proposed method in various practical domains, indicating its significant contribution to the field of gait recognition.

\end{abstract}

%\begin{IEEEkeywords}
\textbf{Keywords:}
Gait recognition, biometrics, person identification, gait landmarks, Procrustes analysis, Siamese biGRU-dualStack Neural Network
%\end{IEEEkeywords}

%\IEEEpeerreviewmaketitle

\section{Introduction}
%\IEEEPARstart{B}
{Biometrics} refers to the automatic identification or authentication of individuals by analyzing their physiological and behavioral characteristics. Physiological biometrics, such as the face, fingerprints, iris, and retina, are stable means of authenticating and identifying people. However, these traits require cooperation from the subject and a controlled environment, making them unsuitable for surveillance systems. Even though these techniques work well in a lot of situations, they can be hard to use in others. They can have problems like obstructed views, and distant or poorly defined data, and frequently necessitate the subject’s cooperation.

Gait recognition identifies individuals based on their walking posture, and is a non-invasive technique that is hard to copy, making it ideal for access control, covert video surveillance, criminal investigation, and forensic analysis. Human walking follows a repeating pattern where the right leg steps, followed by the left leg, and then the right leg again, forming a gait cycle \cite{b1}. This gait cycle encompasses 32 gait features, such as stride, torso movement, hand position, joint angles, foot spacing, and foot length. Gait recognition has the advantage of operating at a distance and with low-resolution images, making it applicable in diverse situations \cite{b2}. However, gait recognition faces challenges related to different intraclass variations in appearance and environment, such as clothing, carrying variation, illumination, walking surface, and view angle, which can significantly reduce performance \cite{b3}.

Gait analysis has been extensively studied, particularly in biometrics and human identification. Researchers have utilized various techniques to extract gait features, including spatiotemporal features, frequency domain features, and wavelet-based features \cite{b4, b5}. These features capture different aspects of gait, such as body segment movements, frequency components, and time-frequency characteristics of gait signals. Physical sensors are commonly used in addition to these techniques to identify gaits \cite{b6}. These sensors are placed on the feet, legs, and torso to measure parameters such as stride length, step time, and cadence. The measurements obtained from these sensors can then be used to extract gait features and classify them using techniques such as SVMs, neural networks, and decision trees \cite{b7, b8}. However, wearable sensor-based approaches have achieved state-of-the-art performances. It is uncomfortable to wear it all day, modeling, and eventually, some people may forget to do so. However, the detective range is constrained by the comparatively expensive installation costs of these environmental-based sensors. 

The field of computer vision has witnessed a surge in the adoption of modern deep learning-based algorithms, which have exhibited exceptional performance in various tasks, including person reidentification, pose estimation, and gait recognition \cite{b9, b10}. These advances have paved the way for significant improvements in gait recognition, a vital aspect of biometric identification. Particularly, advancements in human body pose estimation have proven instrumental in accurately modeling the different body parts necessary for model-based gait recognition. \textcolor{black}{Additionally, Recurrent Neural Network (RNN) \cite{b11}, renowned for capturing long-range dependencies in temporal contexts, have demonstrated promising results in gait recognition tasks \cite{b12}.}

\begin{figure*}[ht]
    \centering
    \includegraphics[width=12.0cm]{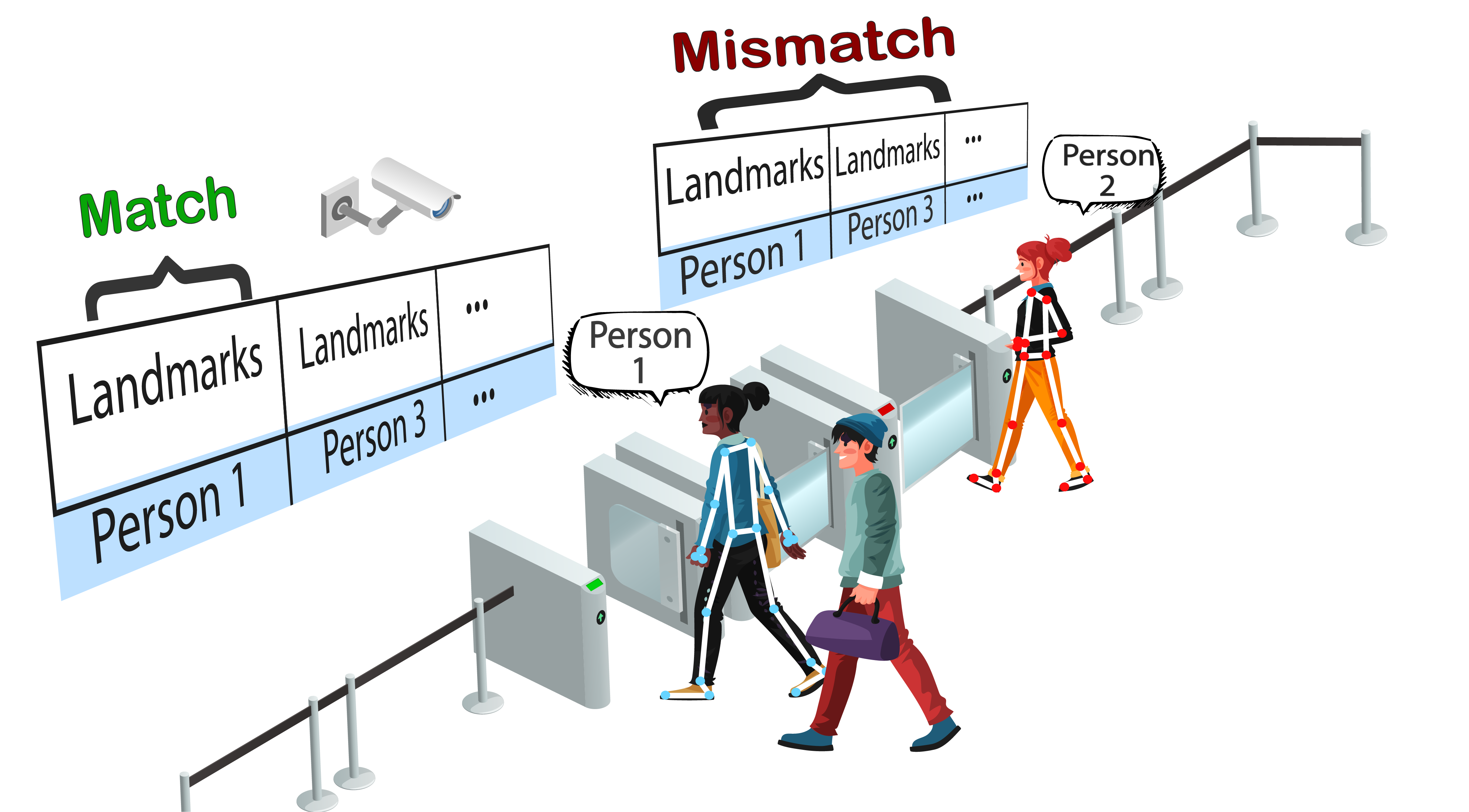}
    \caption{Access Control Based on Gait Sequence Matching: Successful vs. Unsuccessful Cases.}
    \label{overview}
\end{figure*}

\textcolor{black}{Despite the notable advancements achieved in deep learning-based gait recognition, the field still faces several significant challenges \cite{bd1}. A primary obstacle is that spatial-based methods, which offer lower computational costs, can overlook crucial temporal information. Temporal-based methods, while skilled at automatically extracting spatial and temporal features, may miss dynamic frame-to-frame differences that are essential for successful gait recognition \cite{a2, b83, b501}. These computational demands can hinder the practical deployment of deep learning approaches in real-world scenarios.}

To address these challenges, we propose a novel approach that leverages advanced techniques in gait recognition. By utilizing the sequential gait landmarks obtained through the Mediapipe pose estimation model, our approach ensures comprehensive coverage of the gait cycle. We further address the issue of variability caused by different angles of approach by employing Procrustes analysis, which aligns gait frames for enhanced accuracy. \textcolor{black}{To enable dynamic analysis of gait patterns and address the computational challenges, we employ a sophisticated Siamese biGRU-dualStack Neural Network architecture. This design not only captures essential temporal dependencies for comprehensive gait analysis but also streamlines computational complexity, providing an effective solution to manage the inherent computational demands of gait recognition. Our approach has been extensively validated through experiments conducted on large-scale cross-view databases, such as CASIA-B, SZU RGB-D, and OU-MVLP, demonstrating its robustness and reliability in accurately identifying and distinguishing individuals based on their distinctive gait patterns.} In Figure \ref{overview}, a clear distinction is observed between the two individuals depicted. Person 1's gait sequence demonstrates a successful match, granting them access. However, for Person 2, their gait sequence fails to match, resulting in a denied entry.

These promising results underscore the potential applications of our approach across various practical domains, highlighting its significant contribution to the field of gait recognition.

The main contributions of this study are as follows:

\begin{itemize}
  \item Introducing a new approach for gait recognition that accurately identifies and differentiates individuals based on their unique walking patterns.
  \item Using sequential gait landmarks obtained through the Mediapipe pose estimation model to ensure comprehensive coverage of the gait cycle, resulting in a more accurate representation of gait patterns.
  \item Applying Procrustes analysis to align the gait landmarks, minimizing the impact of varying orientations, and improving the accuracy of gait recognition.
  \item Utilizing a Siamese biGRU-dualStack Neural Network architecture with contrastive loss to capture the temporal dependencies in sequential gait data, enabling accurate analysis of gait dynamics and better identification of individuals.
  \textcolor{black}{\item Testing the proposed methods on four significant cross-view datasets: CASIA-B, SZU RGB-D, OUMVLP-Pose, and \textcolor{black}{Gait3D}.}
  
\end{itemize}

The \href{https://github.com/promaprogga/Siamese-BiGRU-dualStack}{\textcolor{red}{code and models}} will be made publicly available soon now avaialble from github: \url{https://github.com/promaprogga/Siamese-BiGRU-dualStack}.

\section{Related work}
Human identification is an important research topic with numerous applications in security, surveillance, and healthcare. Gait analysis \cite{b13, b131} is an attractive method of identification as it can be performed at a distance, is non-invasive, and does not require any special equipment or training. In recent years, there has been growing interest in the use of gait analysis for human identification \cite{b14}, and a considerable body of research has been published on this topic. It started with the advancement of cameras and web technology. However, humans can be identified by biometric traits with high accuracy and reliability due to their unique and distinctive nature. Wearable biometrics is an active research area with interesting applications for real-life scenarios. Current state-of-the-art research has demonstrated that various characteristics can be utilized to accurately identify the users of the employed devices in a continuous manner \textcolor{black}{ \cite{b15, b17, b18}.} 

% \textcolor{red}{
% Many research studies have been conducted on face and fingerprint [19, 20] recognition, which are two of the most widely used biometric identification methods [21].
% }

The accurate identification of individuals from a long distance can be challenging, as the biometric features required for identification may be too small or obscured to be reliably captured. This can be a major obstacle in applications such as security and surveillance, where it is important to be able to identify individuals at a distance. Gait analysis is used to identify individuals from a distance based on their walking patterns \cite{b19}. \textcolor{black}{ Fu et al. \cite{b20} employs a pose-based approach for gait recognition, showcasing comparable results to silhouette-based methods. The proposed GPGait framework introduces HOT, HOD, and PAGCN, demonstrating superior cross-domain performance and potential for effective pose-based gait recognition.}

There are generally two types of gait analysis methods used in the current research community: vision-based, and wearable sensor-based. In the context of signals recorded using video sensors, the Gait Energy Image (GEI) representation has been widely utilized. An improved version of the GEI method is also employed to enhance its effectiveness in \cite{b23}.
\textcolor{black}{Anwary et al. \cite{b24} propose a gait evaluation method using Procrustes and Euclidean distance matrix analysis that collects real-time accelerometer and gyroscope data from inertial measurement unit (IMU) sensors and he investigated the optimal location for wearable sensors. In another study \cite{b28}, the Kinect device is used to capture three-dimensional coordinates of human bones, and the distances between bone nodes are used as features. In addition, Anwary et al. \cite{b25} investigated the optimal location for wearable sensors, automated the extraction of gait parameters, and evaluated gait abnormalities \cite{b27}.} A support vector machine (SVM) classifier is utilized, employing one-versus-one and one-versus-all algorithms to solve the multi-classification task. Another often-used approach is the application of deep Convolutional Neural Networks (CNN). A deep CNN architecture was developed by \cite{b3} consisting of eight layers: four convolution layers and four pooling layers. This architecture is less sensitive to several typical variations and occlusions reducing the quality of gait recognition. Deep CNN has been successfully used to classify images from various sources, as demonstrated in a study \cite{b29}. \textcolor{black}{
Current skeleton-based gait recognition struggles with distinguishing walking styles across views. To address this, Huang et al. \cite{b292} the proposed Condition-Adaptive Graph (CAG) convolution network introduces Joint-Specific Filter Learning (JSFL) and View-Adaptive Topology Learning (VATL) modules. JSFL adapts filters at the joint level, capturing unique patterns, while VATL dynamically adjusts graph topologies based on view conditions.} The study described in \cite{b30} presents the application of the Vision Transformer with an attention mechanism for gait recognition. The gait energy image is computed and splits into patches, which are then embedded and fed into a Transformer for gait representation.

% \textcolor{red}{Vision-based gait recognition is a biometric method that uses video cameras to capture an individual’s walking patterns and extract features for identification purposes [31].} 
\textcolor{black}{Gait recognition systems that do not require individuals to wear any devices predominantly rely on vision and are commonly referred to as vision-based gait recognition. These systems utilize imaging sensors to capture gait data without requiring active cooperation from subjects, even from considerable distances \cite{b12}. There are now two types of conventional gait recognition techniques: appearance-based and model-based techniques. Model-based approaches \cite{m1, b292} mainly rely on the recognition of the human pose structure and movement.}

\textcolor{black}{Model-based gait recognition techniques, such as 2D/3D posture and the Skinned Multi-Person Linear (SMPL) \cite{m3} model, typically use the determined underlying structure of the human body as input. In model-based approaches, researchers use techniques to imitate how the human body moves and the structure of the body by designing simulated models \cite{m4} or incorporating skeletons as inputs \cite{m5, b2}. Specifically, PoseGait \cite{m5} is a model-based approach using 3D human body poses obtained from Convolutional Neural Network estimations. The 3D pose provides invariance to view changes and external factors. Teepe et al. \cite{m8} introduced GaitGraph for leveraging human pose estimation for cleaner gait representations. This approach combines skeleton poses with Graph Convolutional Networks (GCN) for improved spatiotemporal modeling.
} 

Appearance-based approaches \cite{a1} obtain silhouettes as inputs, which rely on abundant shape information to model spatial-temporal features. Some of the representative appearance-based methods are disentanglement-based, set-based, part-based, and 3D convolutional neural networks (CNNs)-based. Here GaitNet \cite{b58}, an end-to-end network integrating silhouette segmentation, feature extraction, learning, and similarity measurement. Comprising two convolutional neural networks for segmentation and classification. Moreover, GaitPart \cite{b83}, an approach that focuses on specific body parts for better gait recognition. It improves performance using the Focal Convolution Layer for detailed spatial learning and the Micro-motion Capture Module (MCM) for short-range temporal features, avoiding unnecessary long-range ones. Chao et al. \cite{a2} introduce GaitSet, a method that learns identity information from gait sets. Operating from a set perspective, GaitSet is immune to frame permutation and seamlessly integrates frames from diverse videos filmed under various scenarios.
{
Pinyoanuntapong et al. introduce GaitMixer \cite{GaitMixer}, a new model for improving skeleton-based gait recognition, addressing the performance gap with appearance-based methods. GaitMixer uses a multi-axial mixer architecture, combining spatial self-attention and temporal large-kernel convolution to capture diverse gait features. This method enhances recognition robustness against changes like clothing and carried items. Tests on the CASIA-B database reveal that GaitMixer surpasses previous skeleton-based techniques and rivals appearance-based approaches. Current gait recognition systems use manual attention mechanisms like cropping silhouettes, limiting their learning capabilities. To overcome this, Castro et al. propose AttenGait \cite{AttenGait}, an approach with trainable attention mechanisms that automatically discover important areas in the input data, achieving state-of-the-art results on the CASIA-B dataset.
}
% \textcolor{red}{
% Within this work, it has been shown that a person can identify another person whom he or she knows by the way that person moves, even when their clothing or hairstyle is not typical, at great distances. Deng et al. [b32] proposed a gait recognition method where Lower limb joint angles are determined as gait parameters and a deterministic learning technique is used to capture the dynamics of the gait system that underlies time-varying gait parameter trajectories. Jahangir et al. [b33] demonstrate a new two-stream deep learning framework for human gait recognition. It includes steps such as contrast enhancement, data augmentation, deep transfer learning, a fusion of features, and an improved feature selection method. Experimental results on the CASIA-B dataset [34] demonstrate improved accuracy compared to state-of-the-art techniques. The framework shows potential for further optimization and analysis of specific angles.
% Deng et al. [b35] proposed a method for recognizing a person’s gait that works from different viewing angles and is based on gait dynamics and knowledge fusion. To get the view-independent features needed for reliable gait recognition, they used RBF neural networks [b36] and a deterministic learning algorithm to get the width parameters underlying the gait system dynamics. They represent the gait patterns through nonlinear dynamics and also create a trained pattern bank.}

In the field of gait recognition, most studies use a CNN to extract the spatial waveform features of gait data \cite{b37}. Some studies use a recurrent neural network (RNN) \cite{b38}, gated recurrent unit (GRU) \cite{b39}, or LSTM \cite{b40} to extract the time-series correlation features of gait data. Khokhlova et al. \cite{b41} propose an LSTM-based model for classifying normal and pathological gait patterns using low-limb flexion angles from the Kinect V2 sensors. Their approach aims to automate gait analysis and provide clinicians with a reliable tool for diagnosing gait-related disorders. By creating 2D CNN, LSTM, and Bi-LSTM models, the authors \cite{b42} made a substantial contribution to the recognition of human activity. The proposed approach \cite{b43}, utilizing radar sensors and Bi-LSTM networks, demonstrates its effectiveness in accurately classifying individual and sequential gaits, including fall events. Low et al. \cite{b44} developed a stacked bidirectional LSTM (Bi-LSTM) model to understand human walking speed Based on kinematic data. Their technique displays the capability to classify various walking speeds by capturing temporal correlations in gait data. Additionally, Albuquerque et al. \cite{b45} presented a framework for pathological gait classification that incorporates a bidirectional LSTM and an optimized VGG-16 CNN \cite{b46}, achieving high accuracy in cross-validation and cross-dataset evaluation and accurate classification of various pathological gaits with robustness to noisy input silhouettes. Cao et al. \cite{b47} developed a framework for predicting the remaining useful life (RUL) of bearings using transfer learning and a bidirectional-GRU (BiGRU) network. Their approach demonstrates the effectiveness of transfer learning in improving RUL prediction accuracy across multiple working conditions. 
% \textcolor{red}{ For ECG-based biometric identification, a deep learning model is proposed by Lynn et al. [48], which incorporates bidirectional training with GRU cells. This model outperforms existing methods by capturing both past and future time steps, improving performance and feature extraction.}
The researchers in \cite{b49} propose a CNN-RNN deep learning model for classifying human emotional states based on human gait data captured by on-body smart devices, achieving high classification accuracies using the 1D magnitude of 3D accelerations as input. The model incorporates dense connections through 1x1 convolutions and combines elements from the InceptionResNet CNN and BiGRU models. Bidirectional GRU performs data processing in both directions, that is in both forward and backward directions and concatenates the resulting output. Stacked Bidirectional GRU increases the depth of the layers used in the GRU model. However, Ullah and Munir \cite{b50} proposed a framework that addresses human activity recognition in video streams through a cascaded spatial-temporal discriminative feature-learning approach. It combines an attentional CNN architecture with a stacked bidirectional gated recurrent unit (Bi-GRU) network, allowing efficient modeling of spatial-temporal dynamics. \textcolor{black}{
There is another model \cite{b19} that includes Global Feature Extractor (GFE) and Dynamic Feature Extractor (DFE) modules, prioritizing spatial-temporal and dynamic features, respectively.
Lin et al. \cite{b501} introduce a Global and Local Feature Extractor (GLFE) employing multiple global and local convolutional layers (GLConv). Additionally, they present Local Temporal Aggregation (LTA), an approach to enhance spatial resolution by reducing temporal resolution.}
The stacked Bi-GRU captures long-term temporal dependencies using forward and backward gradient learning, utilizing knowledge from both previous and upcoming frames \cite{b51}.
\textcolor{black}{Huang et al. \cite{b511} proposed Context-Sensitive Temporal Feature Learning (CSTL) network addresses challenges in learning discriminative temporal representations by aggregating temporal features across multiple scales, considering temporal relations, and addressing misalignment problems by providing a Salient Spatial Feature Learning (SSFL) module. Lin et al. \cite{b512} introduce a Multi-scale Temporal Feature Extractor, capturing both the subtle and swift changes in gait to address approaches using 3D CNNs that tend to miss out on details by focusing solely on one temporal scale.}

However, Mediapipe \cite{b52} developed by Google, stands as a powerful and highly effective tool for gait recognition in the field of biometrics. It is an innovative open-source project that offers a comprehensive, yet streamlined, solution that encompasses speed, simplicity, cost-effectiveness, portability, and ease of deployment. This remarkable framework empowers developers to construct applied machine-learning pipelines capable of handling various types of data, including video, audio, and time-series information. One of the key strengths of Mediapipe lies in its advanced pose estimation \cite{b53} capabilities, making it particularly well-suited for gait analysis. These include gesture recognition \cite{b54}, hand landmarks, image classification, object detection, and face landmarks, among others. Kim et al. \cite{b55} employed MediaPipe to estimate 2D human joint coordinates in each image frame. They utilize the BlazePose architecture, which extracts 33 two-dimensional human body landmarks. The authors then presented a 3D human pose estimation system that takes the 2D skeletal poses estimated by MediaPipe as input and fits them to a 3D humanoid robot model using an optimization method called uDEAS. Experimental validation shows acceptable accuracy and suggests potential applications in activity recognition and the analysis of construction workers and patients with Parkinson’s disease. \textcolor{black}{ Moreover, the re-extraction of landmarks using the MediaPipe pose estimation technique in the study conducted by Garg et al. \cite{b551} serves a specific purpose within their proposed 3D human pose estimation system. While a publicly available pose model dataset already exists, the authors opted for the re-extraction of landmarks using MediaPipe Pose to overcome certain limitations associated with deep learning methods. Deep learning models often face challenges in accurately estimating poses that are absent or rare in their training datasets. By utilizing the off-the-shelf 2D pose estimation method, MediaPipe Pose, the authors obtain 2D skeletal poses from monocular images. This approach provides a lightweight alternative, allowing for the estimation of joint angles for 3D pose without the computational demands of high-performance PCs or GPUs. Additionally, the MediaPipe Pose technique aids in addressing depth ambiguity issues in 3D pose estimation. The re-extraction of landmarks using MediaPipe Pose, combined with an optimized 3D humanoid robot model, contributes to the overall effectiveness and real-time feasibility of the proposed pose estimation system, making it suitable for applications in mobile robot systems.}

In recent years, Siamese networks \cite{b56} have emerged as a promising solution for tackling the difficulties associated with gait recognition. This approach has gained attraction due to its ability to address key challenges, including the limited number of instances for each subject and the domain disparity between gait sequences and traditional image classification tasks. Researchers have proposed several innovative approaches that leverage Siamese networks to improve gait recognition performance \cite{b57}. For instance, Songa et al. \cite{b58} have proposed GaitNet as a way to learn segmentation and recognition of gait at the same time. It is an end-to-end pipeline and can automatically discover discriminative information for gait recognition. Two convolutional neural networks comprise it: one for classification and the other for gait segmentation.
% \textcolor{red}{A single joint learning procedure can be used to train the two networks at the same time.}
 Similarly, Zhang et al. \cite{b59} proposed a Siamese neural network for gait recognition that utilizes Gait Energy Images (GEIs) as a substitute for raw gait sequences. GEIs filter out extraneous data while retaining key human figures and gait variations, enabling Siamese networks to efficiently extract distinctive biometric information. By employing a distance metric learning architecture that minimizes the distance between similar subjects and maximizes the distance between dissimilar pairs, combined with the use of the K-Nearest Neighbor (KNN) algorithm for human identification in surveillance settings. They achieved significant advancements in gait recognition accuracy. 
% \textcolor{red}{In the domain of human re-identification in surveillance systems, Varior et al. [60] introduced a Siamese Convolutional Neural Network (S-CNN) architecture. S-CNN captures local patterns but overcomes limitations in accurately distinguishing between positive pairs and hard negatives by incorporating a matching gate. This gate enhanced feature extraction by promoting local similarities across higher layers of the network. The proposed methodology has exhibited greater accuracy in human re-identification compared to other deep learning architectures. To further improve gait recognition performance,}
Liu et al. \cite{b61} proposed a comprehensive framework that utilizes competitive gait energy images (GEI) and Convolutional 3D 
\begin{figure*}[ht]
    \centering
    \includegraphics[width=15.5cm, height=4.5cm]{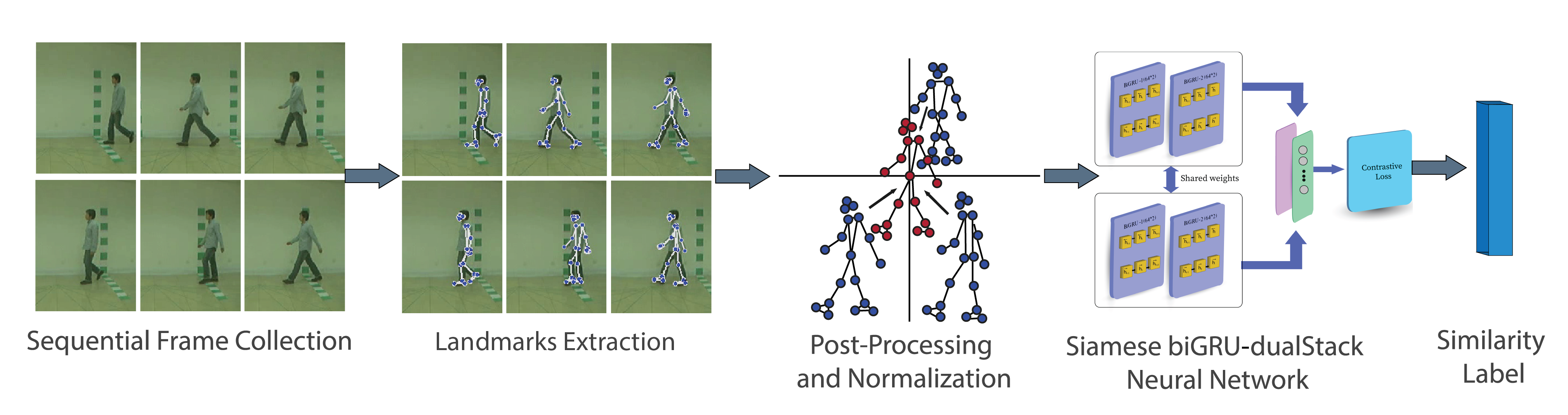}
    \caption{Proposed Framework for Gait Recognition Using Sequential Landmarks.}
    \label{fig1}
\end{figure*}
(C3D) presentations as network inputs. By using a Siamese neural network to directly calculate the resemblance between two human gaits and incorporating Null Space Fractional Transform (NSFT) to merge GEI and C3D characteristics, they achieved more robust and discriminative spatial-temporal gait features, outperforming existing state-of-the-art techniques. Additionally, Bedi et al. \cite{b62} presented an end-to-end LSTM-VGRNet2 network for gait recognition. This network utilizes a novel representation of gait video frames known as stereo silhouette maps. By employing a 3D Convolutional Neural Network (CNN) model for extracting spatio-temporal features and an LSTM network for effectively learning inter-GCS variation.
\textcolor{black}{Gait recognition methods still have problems with adaptability to varying viewpoints and individual appearances and often struggle to capture fine-grained spatio-temporal features. Spatio-Temporal Augmented Relation Network \cite{b63} adaptively generates salient features in diverse regions for mining and extracts spatio-temporal augmented features with accurate temporal scales.}

The training process incorporates hard negative mining and dynamic adaptive margin techniques, resulting in improved performance on challenging datasets such as CASIA-B and OU-ISIR \cite{b64} Gait. Siamese Recurrent Networks (SRN) \cite{b65} has also been employed to enhance the precision of gait recognition systems by leveraging their ability to process time series data. Wang et al. \cite{b66} proposed a novel gait recognition method based on a Conv-LSTM network model that takes advantage of the inherent temporality of human gait. Through comprehensive comparisons and analysis of CASIA-B and OU-ISIR datasets, the proposed method demonstrated superior performance compared to existing approaches, significantly improving recognition rates. The authors of these studies have emphasized the advantages of gait recognition over traditional biometrics, such as face and fingerprint, and have discussed the challenges in gait recognition, including cross-view variations, different clothing, multiple carrying conditions, and low image resolution.

\section{Methodology}
In this research, we propose a method for gait recognition using sequential gait landmarks. The primary objective of this approach is to accurately identify and distinguish individuals based on their gait patterns.

Then, the pose estimation model (MediaPipe) is used to capture sequential gait frames (N) based on foot landmarks, ensuring that frames complete a full walking cycle and the corresponding landmarks of individuals are collected. To address the variability caused by individuals approaching from different angles, we applied Procrustes analysis to align the gait frames. Finally, we employ a Siamese BiGRU-dualStack Neural Network to identify individuals based on gait, as shown in Fig. \ref{fig1}. The Siamese BiGRU-dualStack Network takes pairs of gait sequences as input and learns to distinguish between different individuals. Through experiments and evaluations, we demonstrate the effectiveness of
\begin{figure*}[ht]
    \centering
    \includegraphics[width=14cm]{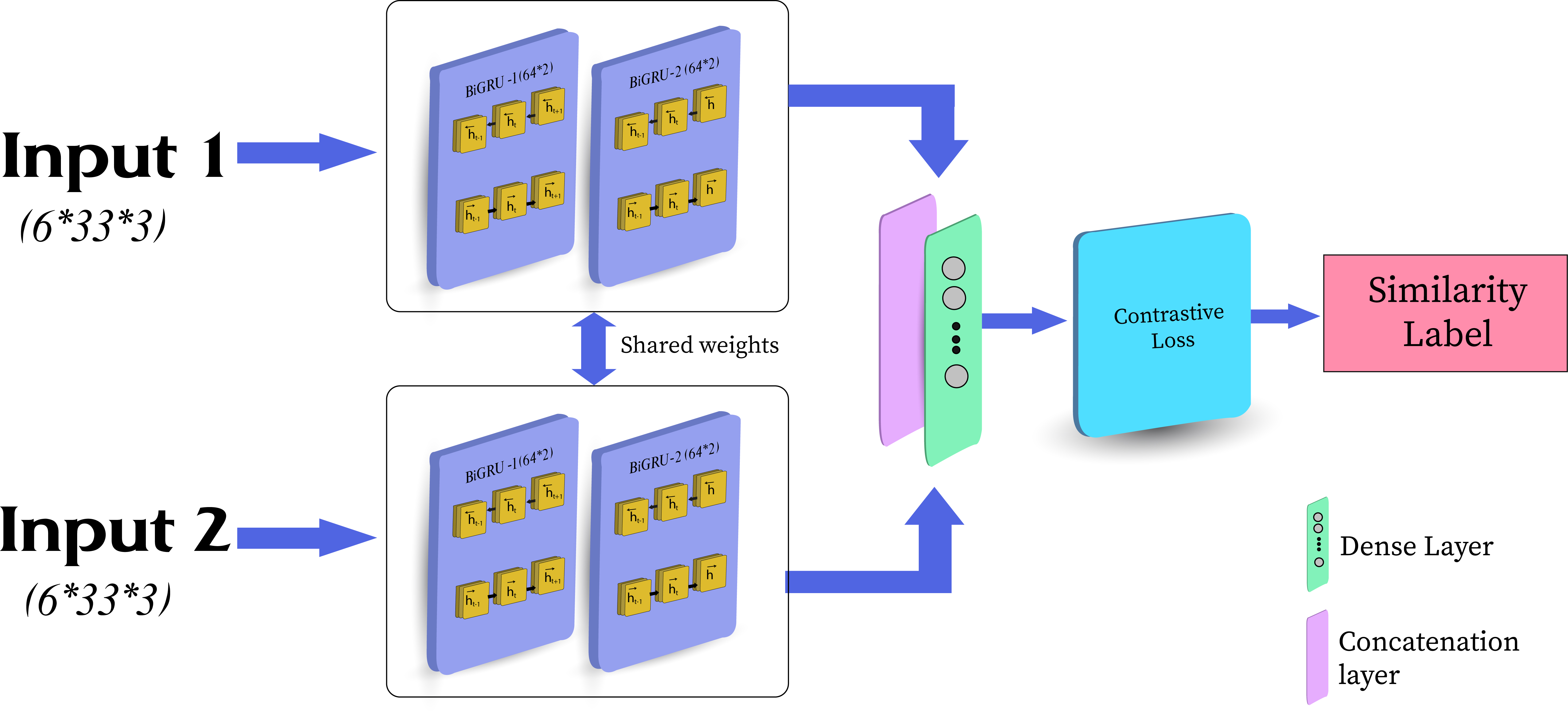}
    \caption{Siamese BiGRU-dualStack Neural Network Architecture.}
    \label{fig2}
\end{figure*}
our proposed approach to accurately identifying individuals based on their gait patterns.

Recurrent Neural Networks (RNNs) are a subset of neural networks that is specifically intended to process sequential data by storing information from previous time steps in hidden states. However, traditional RNNs suffer from the vanishing gradient problem, which occurs during training when gradients diminish exponentially as they propagate through time. The RNN's capacity to identify long-term dependencies in sequential data is hampered by this problem. To address the vanishing gradient problem and capture long-term dependencies more effectively, the Gated Recurrent Unit (GRU) \cite{b67} and other recurrent neural network variations, such as the Long Short-Term Memory (LSTM) architecture, were introduced.

GRUs are specifically designed to address the vanishing gradient problem in traditional RNNs, which occurs when gradients diminish exponentially over time and make it difficult for the network to learn long-term dependencies. GRUs achieve this by using gating mechanisms that allow the network to selectively update and reset information. While GRUs excel at capturing long-term dependencies, they also perform well at modeling short-term dependencies. GRUs typically have fewer parameters than standard RNNs. Traditional RNNs have separate input, output, and hidden state parameters, whereas GRUs combine these into a single update gate and reset gate. This reduction in parameters makes GRUs more memory-efficient and computationally faster. Moreover, having fewer parameters reduces the risk of overfitting, especially when working with limited training data. GRU has two main gates: an update gate and a reset gate, which control the flow of information through the network. The previous hidden state (h) is divided into two parts: the amount that should be sent to the current time step (t) by the update gate (z) and the amount that should be forgotten by the reset gate (r).

However, bidirectional GRU (BiGRU) \cite{b48} is an extension of the GRU model that incorporates information from both past and future time steps. It addresses the limitations of traditional GRU and allows the model to capture dependencies in both directions. In a BiGRU, the input sequence is processed in two directions: forward and backward. The forward GRU processes the sequence from the beginning to the end, while the backward GRU processes it from the end to the beginning. The output of the BiGRU is obtained by concatenating the hidden states from both the forward and backward GRU layers. This combined representation captures the context of each time step in the sequence.
The formulas for computing the hidden states in a BiGRU are as follows:

Forward GRU:
\vspace{-0.5em}
\begin{equation} \label{eq2}
z_{f(t)} = \sigma(W_{z_f} * [h_{f(t-1)}, x_t])
\end{equation}
\begin{equation} \label{eq3}
r_{f(t)} = \sigma(W_{r_f} * [h_{f(t-1)}, x_t])
\end{equation}
\begin{equation} \label{eq4}
h~_{f(t)} = \tanh(W_f * [r_{f(t)} * h_{f(t-1)}, x_t])
\end{equation}
\begin{equation} \label{eq5}
h_{f(t)} = (1 - z_{f(t)}) * h_{f(t-1)} + z_{f(t)} * h~_{f(t)}
\end{equation}

Backward GRU:
\vspace{-0.5em}
\begin{equation} \label{eq6}
 z_{b(t)} = \sigma(W_{z_b} * [h_{b(t+1)}, x_t])
\end{equation}
\begin{equation} \label{eq7}
 r_{b(t)} = \sigma(W_{r_b} * [h_{b(t+1)}, x_t])
\end{equation}
\begin{equation} \label{eq8}
h~_{b(t)} = \tanh(W_b * [r_{b(t)} * h_{b(t+1)}, x_t])
\end{equation}
\begin{equation} \label{eq9}
h_{b(t)} = (1 - z_{b(t)}) * h_{b(t+1)} + z_{b(t)} * h~_{b(t)}
\end{equation}

Combined Output:
\begin{equation} \label{eq10}
h_t = [h_{f(t)}, h_{b(t)}]
\end{equation}

In the above formulas, $z_{f(t)}$ and $z_{b(t)}$ are the update gate activations for the forward and backward GRUs, respectively. On the other hand, $r_{f(t)}$ and $r_{b(t)}$ are the reset gate activations, $h~_{f(t)}$ and $h~_{b(t)}$ are the candidate hidden states and $h_{f(t)}$ and $h_{b(t)}$ are the forward and backward hidden states at time step $t$.

% The Siamese network consists of two identical subnetworks, or branches, which share the same weights and architecture. Each branch processes a separate input sequence, allowing for direct comparison between them. In this case, each branch of the Siamese network incorporates two bidirectional Gated Recurrent Units (GRUs) with 128 units and a Rectified Linear Unit (ReLU) activation function. Fig. \ref{fig2} shows an overview of the network architecture. Two bidirectional GRUs in each branch of the Siamese network can capture and encode the temporal dynamics of the gait sequences from both directions. This bidirectional analysis provides a more comprehensive representation of the gait patterns, improving the network's ability to recognize and distinguish between different individuals based on their gait. However, after processing the input sequences in the bidirectional GRUs, the outputs of both branches are concatenated. The concatenated output is then passed through a 1x1 dense layer. This layer performs a linear transformation on the concatenated features, followed by a sigmoid activation function. 
\textcolor{black}{The proposed model incorporates a Siamese network, a specialized architecture comprising two identical branches that share weights and structures. This enables direct comparison between input sequences, enhancing the model's capability to discern nuanced gait patterns. In this case, each branch of the Siamese network incorporates two bidirectional bidirectional Gated Recurrent Units (GRUs) with 128 units and a Rectified Linear Unit (ReLU) activation function. This configuration is designed to capture and encode the temporal dynamics of the gait sequences effectively. In each branch, the bidirectional GRUs enable the network to analyze the gait information in both forward and backward directions, allowing a comprehensive understanding of the gait patterns within the sequences. This bidirectional approach enhances the network's ability to recognize and distinguish between different individuals based on their gait. However, after processing the input sequences in the Bidirectional GRUs, the outputs of both branches are concatenated. Fig \ref{fig2} shows an overview of the network architecture. Following the concatenation, a 1x1 dense layer is applied to the combined output. This dense layer serves as a transformation step, linearly mapping the concatenated features to a new space. The use of a 1x1 dense layer allows for a flexible adjustment of the feature dimensions. Subsequently, a sigmoid activation function is applied to the transformed features. The sigmoid activation function is chosen to introduce non-linearity and ensure that the network produces output values within the range of [0, 1].  
}

Overall, the approach aims to advance the field of gait recognition by providing an effective methodology for gait feature extraction, comparison, and identification. The experimental results and evaluation demonstrate its efficacy in accurately recognizing and distinguishing individuals based on their unique gait characteristics.

\section{Experimental Analysis}
\subsection{Dataset}
% Two datasets, CASIA-B and SZU RGB-D, are involved in our experiments to evaluate the proposed method. 
We conducted extensive evaluations of our proposed model using indoor datasets such as CASIA-B, SZU, and OU-MVLP. \textcolor{black}{Furthermore, to enrich our analysis, we incorporated the Gait3D dataset.} Each dataset offers unique characteristics and challenges, contributing to a comprehensive assessment of our model's performance.

\begin{figure}[ht]
    \centering
    \includegraphics[width=8cm]{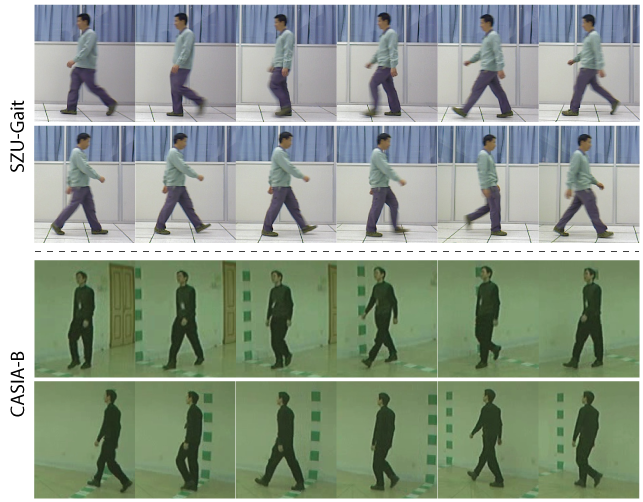}
    \caption{Overview of CASIA-B and SZU RGB-D Gait Datasets.}
    \label{fig4}
\end{figure}

CASIA-B \cite{b34}: CASIA-B gait dataset stands out as one of the most extensive publicly available repositories of gait information. The images in the CASIA-B gait dataset are stored in PNG format, and each image has a resolution of 128 X 64 pixels. The labels contain information about the subject ID. It includes 124 subjects, captured from 11 viewpoints (93 men and 31 women). The view range is $0^{\circ}$ to $180^{\circ}$, with a distance of $18^{\circ}$ between the two closest perspectives. There are 6 normal walking sequences ("nm"), 2 bag walking sequences ("bg"), and 2 coat walking sequences ("cl"). Figure \ref{fig4} shows the samples from different views of a subject's normal walking. In line with the experimental setup of previous studies \cite{b68, b69, b70}, we adopt a similar approach for subject partitioning. Specifically, we allocate the first 74 subjects for training purposes, while the remaining subjects are reserved for testing.

SZU \cite{b71}: SZU is a large RGB-D gait dataset. It contains 99 subjects, with 8 sequences for each subject in two different views. The first one is the side view $90^{\circ}$, and the second is about $30^{\circ}$ away from the side view $60^{\circ}$. For each view, there were 4 video sequences captured. Two sequences are right-walking ones, and two are left-walking. So there are 8 different sequences for each subject. When subjects walk, synthesized color images (RGB images) and depth images are captured. Gait data from 99 subjects was stored in 792 (99 × 4 × 2views) sequences. Figure \ref{fig4} illustrates various walking motions, captured from different angles. The color and depth image resolutions are all 640 X 480 and are stored in PNG format. Following the experimental setup described in \cite{b72}, we assigned the first 49 subjects for training, while the remaining subjects were reserved for testing.

\textcolor{black}{OU-MVLP \cite{b721}: Multi-View Large Population Database with Pose Sequence (OUMVLP-Pose), is one of the largest gait datasets and comprises 10,307 subjects, each having two sequences. The viewpoints are evenly distributed across [0°, 90°] and [180°, 270°]. Adhering to the established protocol, the first sequence for each ID serves as the gallery, and the subsequent sequences function as probes during the evaluation. In detail, the dataset consists of 28 sequences with 14 camera views per subject, resulting in two sequences ('01' and '02') per view. The initial 5,153 subjects are utilized for training, while the remaining 5,154 subjects are allocated for testing. During testing, the sequences indexed as '01' are designated as the gallery, while those indexed as '02' constitute the probe set.}

\textcolor{black}{Gait3D \cite{gait3d}: A large-scale dataset for gait recognition that focuses on dense 3D representations. It consists of data from 4,000 subjects and over 25,000 sequences, captured from 39 cameras in unconstrained indoor environments. The dataset includes 3,000 subjects for training and 1,000 for testing. It also features 3D Skinned Multi-Person Linear (SMPL) models reconstructed from video frames, offering rich 3D information on body shape, viewpoint, and dynamics.
}
\subsection{Data Pre-processing}
Data preprocessing involves three essential steps: sequential frame extraction, landmark collection, and normalization. These steps ensure the input data is appropriately prepared for subsequent analysis and model training.

Mediapipe is a powerful framework developed by Google that provides a comprehensive solution for various multimedia processing tasks, including pose estimation. It provides a set of pre-built tools and models that make visual data analysis accurate and efficient. In the context of this work, Mediapipe plays a significant role in data preprocessing by performing sequential frame extraction and landmark collection. The Mediapipe Pose estimation model specifically focuses on the accurate detection and localization of human poses, capturing key points and landmarks that define the body's posture and configuration.

The first step of data preprocessing \textcolor{black}{for two datasets such as CASIA-B, SZU} involves sequential frame $(N)$ extraction based on the foot landmark. In this case, the left foot landmark is utilized. By leveraging the Mediapipe Pose algorithm \cite{b55}, the sequential frames $(N=6)$ are extracted in a manner that ensures the left foot landmark values progress from the most negative to the most positive. This sequential arrangement effectively represents the complete cycle of a walking motion. However, in our study, we consider six sequential frames following the experimental setup described in \cite{b58}, as it has been found to provide a satisfactory level of accuracy. By focusing on the left foot landmark and extracting frames in this manner, the resulting sequential frames encapsulate the relevant temporal information necessary for gait recognition and analysis. 

Then, the Mediapipe Pose estimation model is utilized to collect landmarks from each frame. For each individual, their gait data comprises a sequence of six frames. In each frame, the Mediapipe Pose estimation model collects 33 landmarks, with each landmark represented by three coordinates $(x, y, z)$. As a result, the total number of values collected for a single individual amounts to $33 * 3 * 6$, which equals $594$ values. The $x$, $y$, and $z$ coordinates of each landmark represent the spatial positioning of specific key points in the gait sequence. These key points correspond to various body parts, joints, or limbs, providing detailed information about the posture and configuration of the individual during each frame. By collecting these landmarks from all frames, a comprehensive representation of the gait sequence is obtained. The sequential arrangement of the landmarks preserves the temporal dynamics of the gait, while the $x$, $y$, and $z$ coordinates capture the three-dimensional spatial information. \textcolor{black}{The OUMVLP-Pose dataset organizes each sample as a sequence of frames. Within each frame, a set of pose points are recorded, resulting in a comprehensive representation of an individual's pose. The cumulative values captured within each frame across the entire sequence for a single person form the basis for analyzing the data and training the model.}
\textcolor{black}{In the Gait3D dataset, sequences are sourced from 4,000 subjects, with 3,000 subjects used for training and 1,000 for testing.
}

\subsection{Procrustes Analysis}

Normalizing the lengths and times of gait features is a common method for quantifying and comparing human walking patterns. These features encompass eight distinct aspects, including stride length, stride time, stride rate, step length, step time, step speed, stance time, and swing time. These measurements are taken from Cartesian coordinates representing the movements of the right and left legs. The x and y axes represent the characteristics of the respective right and left legs, while dimensionless values unify the data. This framework facilitates the visualization of how both legs move through the depiction of feature curves. Procrustes analysis \cite{b73} is employed to examine shape variations within a dataset. It is a mathematical and statistical approach that disregards time and size when assessing curve shape and shape changes. In this context, the Ordinary Procrustes Analysis (OPA) finds the optimal translation vector, rotation matrix, and scaling factor to align two configurations closely. Generalized Procrustes Analysis (GPA) is utilized to find the best-fit model within a group of entities \cite{b75}. This method avoids the necessity of comparing all potential matrix pairs separately. Instead, it simplifies the process by uniformly adjusting rotation, translation, and scale to achieve the best possible fit. GPA is particularly advantageous for investigating Normalized Mean Gait Shapes (NMGS) and studying the walking patterns of individuals. It emerges when multiple data matrices exhibit a least squares relationship.

Consider a set of m matrices denoted as $X_i (i = 1, 2, 3,..., m)$ representing configurations with landmarks indicating gait traits. These landmarks are described by k shapes, with variations in size or shape. Changes in translation, rotation, and size of a configuration are denoted by $c_i$ (scale factor), $O_i$ (rotation matrix), and $t$ (translation vector) respectively. The relationship is described as:

\begin{equation} \label{eq111}
\hat{X}_i = c_iX_iO_i + jt_{i}^{T} 
\end{equation}

Here, $X$ represents the new point locations of interest in the configuration. The objective of GPA is to transform, rotate, and scale configurations iteratively to minimize the sum of squared distances between corresponding points, thus achieving the best possible alignment among configurations. Iterative steps within the GPA process aim to minimize discrepancies. The shapes undergo resizing, rotation, and translation adjustments until the sum of squared distances reaches a predefined threshold. This process results in a reduction of similar features across all shapes. With a focus on gait traits, Procrustes superimposition determines a representative shape, termed Normalized Mean Gait Shape (NMGS), for individuals. This analysis excludes scaling and reflection operations. The Procrustes diagram visually demonstrates individual walking patterns by highlighting residuals, which indicate differences between landmarks and the NMGS.
\begin{figure}[ht]
    \centering
    \includegraphics[width=8cm]{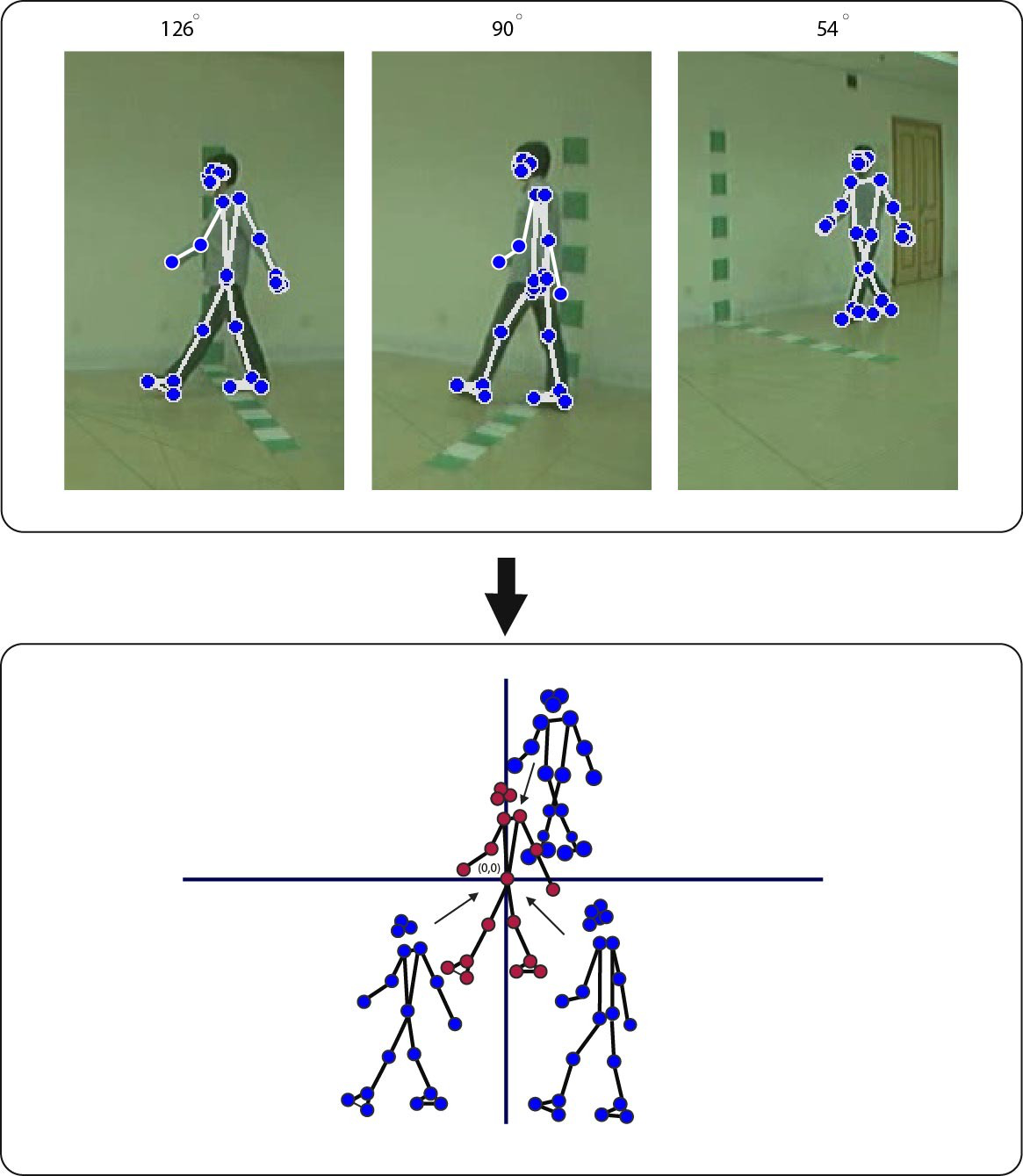}
    \caption{Landmark Shape Alignment Using Procrustes Analysis.}
    \label{fig5}
\end{figure}
Since a person can approach from different angles, Procrustes analysis is utilized to align the gait sequences and ensure consistent spatial positioning across different individuals. Procrustes analysis is a statistical method that adjusts the position, scale, and orientation of a set of points to minimize the differences between them. Hence, the alignment process involves scaling, rotation, and translation adjustments to achieve the best possible alignment between the landmarks of different individuals. Figure \ref{fig5} illustrates the process in which the distinct configurations are adjusted through translation, rotation, and scaling to align with each other. This alignment aims to attain the optimal fit among the individuals. It helps to mitigate the effects of differences in starting positions, camera angles, and body orientations, enabling a more reliable comparison and analysis of gait patterns. 

\subsection{Model Training}
\begin{figure}[ht]
    \centering
    \includegraphics[width=9cm]{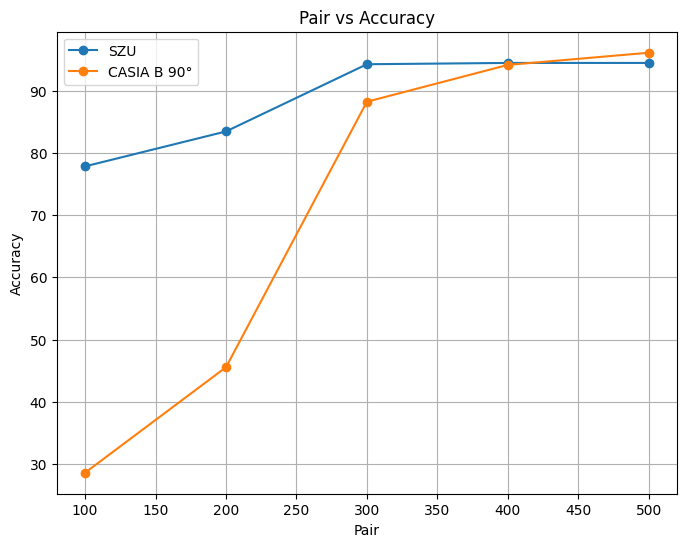}
    \vspace{-1em} 
    \caption{Relations Between Dataset Pair Size and Accuracy.}
    \label{fig6}
\end{figure}

During the model training phase, pairs were constructed to facilitate the learning process. Positive pairs were created using gait sequences from the same individual, while negative pairs were formed by pairing gait sequences from different individuals. For the CASIA-B dataset, which consists of 124 individuals, 74 individuals were used for training purposes. Since there are 74 individuals in the training set, the total number of positive pairs is also 74. However, for negative pairs, we need to consider the combinations of individuals. The number of negative pairs can be calculated as $74C2$. This yields a larger number of possible negative pairs. To ensure a diverse and representative set of negative pairs, we randomly selected a subset of negative pairs from the total number of possible combinations. Figure \ref{fig6} demonstrates the relationship between the number of pairs and the performance of the model. As the number of pairs increases, the model's performance improves. Based on this observation, we chose to utilize 400 pairs for training. in both the CASIA-B and SZU datasets. In the CASIA-B dataset, 74 pairs are considered positive, indicating similar samples, while the remaining (326) pairs are labeled as negative, representing dissimilar samples. On the other hand, in the SZU dataset, 49 pairs are labeled as positive, indicating similar samples, while the remaining (351) pairs are considered negative, representing dissimilar samples. \textcolor{black}{Similarly, for the OU-MVLP Gait dataset, the training phase involved the systematic creation of pairs. A 1:1 ratio of positive and negative pairs was maintained to ensure a diverse and representative training set. \textcolor{black}{However, for the Gait3D dataset, characterized by its wild and diverse nature, we maintained a 1:2 ratio of positive and negative pairs to enhance learning performance. This approach uses 3,000 positive samples and the remaining subjects as negative samples to ensure that the models are well-equipped to handle a variety of scenarios and differentiate between individuals effectively.
}}

During the training process, the model was optimized using the ADAM optimizer with a learning rate of $0.0001$. A batch size of 32 was used, and the training process was repeated for a total of 10 epochs.

\subsection{Evaluation Metrics}
To assess the performance of our proposed approach, we employed several evaluation metrics, including Contrastive Loss and Accuracy. These metrics provide insights into the effectiveness and accuracy of our gait recognition system. 

Contrastive Loss \cite{b76} is a commonly used loss function in siamese network-based models for gait recognition. It measures the similarity or dissimilarity between pairs of gait sequences. The contrastive loss encourages similar gait sequences to have a smaller distance or dissimilarity score, while dissimilar sequences are encouraged to have a larger distance. By minimizing the contrastive loss, we aim to enhance the discrimination and separability of gait patterns. The contrastive loss is computed using the distance or dissimilarity metric between the feature representations of paired gait sequences. The contrastive loss (L) is calculated using the Euclidean distance metric and is defined as follows:
\begin{equation} \label{eq11}
 L = (1 - Y) \;*\; D^2 \;+\; Y \;*\; max(0,\; m - D)^2
\end{equation}
where:
$Y$ is the binary label indicating whether the pair of gait sequences is similar ($0$ for similar, $1$ for dissimilar). $D$ is the Euclidean distance between the feature representations of the paired gait sequences. $m$ is a hyperparameter that controls the separation margin between similar and dissimilar pairs. This loss function helps to optimize the model parameters and improve the overall accuracy of gait recognition.

Rank 1 accuracy is a specific evaluation metric commonly used in gait recognition research to assess the performance of a system in correctly identifying an individual from a gallery of candidates based on their gait patterns. It measures the accuracy of the top-ranked prediction, considering only the most probable match. Rank 1 accuracy can be calculated as follows:\\
\textit{Rank 1 Accuracy} 
\begin{equation}\label{acc}
= \frac{\textit{Number of correctly identified individuals at rank 1}} {\textit{Total number of individuals}}*100
\end{equation}

This metric focuses on the top-ranked prediction, indicating the system's ability to correctly match an individual's gait pattern to their identity among all the candidates in the gallery. By employing the rank 1 accuracy metric, we can specifically evaluate the system's performance in identifying individuals accurately, without considering lower-ranked predictions. It provides a measure of the system's effectiveness in the most critical scenario, where the highest confidence match is expected to be the correct one.

The combination of both contrastive loss and rank 1 accuracy metrics allows us to comprehensively evaluate the proposed approach. While the contrastive loss assesses the optimization aspect and the model's ability to learn discriminative features, the rank 1 accuracy provides a focused evaluation of the system's performance in correctly identifying individuals at the top rank. By considering both metrics, we gain insights into both the optimization and recognition performance aspects of our gait recognition system.

\subsection{Individual Gait Differences}

In the context of this study, we have examined the latent space representation of gait patterns using encoded vectors generated by a computational model. The encoded vectors effectively capture the complex intricacies of gait dynamics, providing a robust method for assessing the similarities and variations in individuals' gait patterns.

In order to measure the magnitude of these disparities, we have utilized the Euclidean distance metric. The calculation of the Euclidean distance between two vectors offers a direct and uncomplicated method for quantifying their dissimilarity inside a multi-dimensional space. The method considers the magnitude of variations along each dimension and calculates the Euclidean distance between the ends of the two vectors. A higher Euclidean distance observed in the encoded vectors indicating gait patterns indicates a more pronounced disparity in the gait dynamics among the individuals.

\begin{figure}[ht]
    \centering
    \includegraphics[width=8cm]{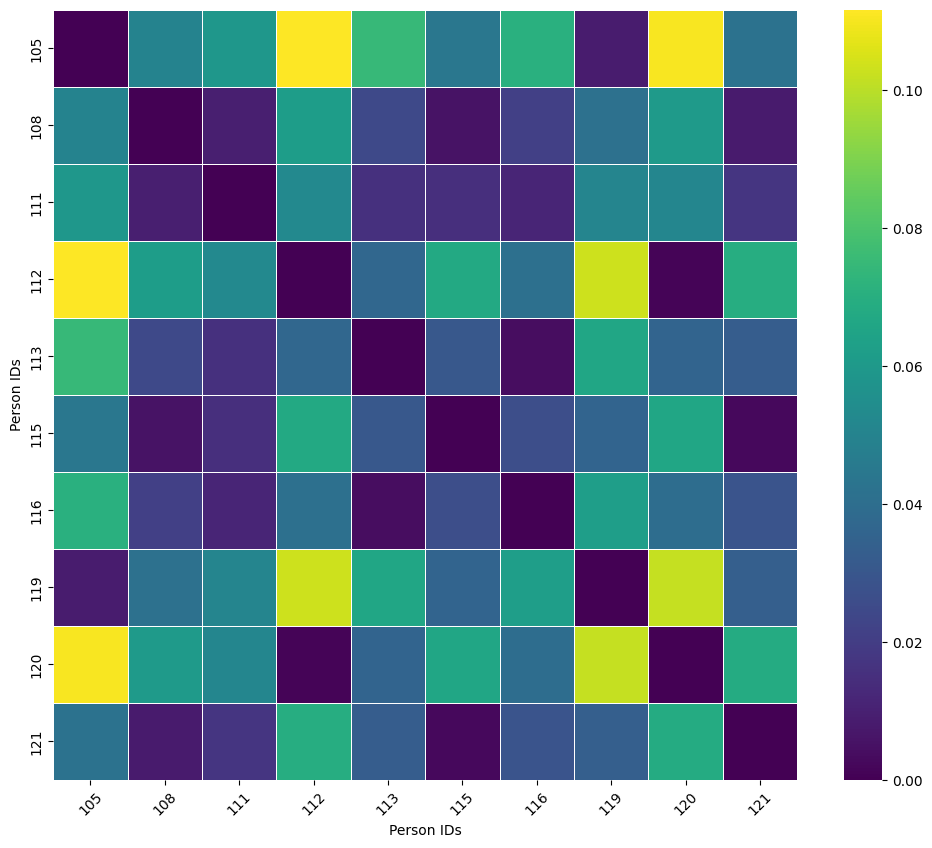}
    \caption{Visualizing Encoded Gait Variations for a Subset of Individuals.}
    \label{fig8}
\end{figure}

\begin{table}[ht]
\centering
\renewcommand{\arraystretch}{1.4}
\caption{Comparative Euclidean Distances in Gait Patterns Among Individuals.}
\setlength{\tabcolsep}{5pt}
\rowcolors{2}{white}{gray!20}
\begin{tabular}{|c|c c c c c c c c|}
\hline
   &  P1   &  P3   &  P4   &  P5   &  P8   &  P9   &  P12   &  P13   \\
\hline
 P1 & 0.000 &  &  &  & &  & & \\
 P3 & 0.002 & 0.000 &  &  &  & &  &  \\
 P4 & 0.075 & 0.077 & 0.000 &  &  &  &  &  \\
 P5 & 0.014 & 0.015 & 0.062 & 0.000 &  &  &  &  \\
 P8 & 0.096 & 0.097 & 0.021 & 0.082 & 0.000 &  &  &  \\
 P9 & 0.123 & 0.124 & 0.048 & 0.109 & 0.027 & 0.000 & &  \\
P12 & 0.042 & 0.041 & 0.117 & 0.056 & 0.138 & 0.165 & 0.000 &  \\
P13 & 0.049 & 0.050 & 0.027 & 0.035 & 0.047 & 0.074 & 0.091 &  0.000 \\
\hline
\end{tabular}
\label{tab:mytable}
\end{table}
Table \ref{tab:mytable} presents a comprehensive quantitative analysis of the Euclidean distances across various individuals' gait patterns. This facilitates an in-depth analysis of the data, which is organized in an 8 × 8 symmetric matrix. The results of our study provide novel insights regarding the unique characteristics of gait patterns. When examining the walking patterns of a single individual, it is constantly observed that the distances are almost zero. The observed result confirms the model's capacity to effectively capture the fundamental regularity that characterizes an individual's walking pattern, as expected. In contrast, significant variations in Euclidean distances are observed when comparing the walking patterns of various individuals. Greater values are suggestive of significant variations from the fundamental gait patterns, so suggesting the existence of distinct gait dynamics among individuals. Figure \ref{fig8} visually represents the variation in encoded gait vectors among random 15 individuals. The heatmap utilizes color gradients to highlight the magnitude of variations, with stronger colors representing more significant disparities. 

\subsection{Comparison with the SOTA Methods}

\textcolor{black}{
In this section, we present a comprehensive analysis of the empirical results obtained from our experiments, focusing on a meticulous evaluation of our method's performance. Our assessment covers the utilization of four distinct datasets, enabling a thorough comparative analysis against existing models. For the CASIA-B dataset, our evaluation encompasses model-based gait recognition methods and also outperforms several state-of-the-art approaches. In the appearance-based category, we consider contemporary models such as VTM \cite{b77}, ViDP \cite{b78}, LRDF \cite{b70}, C3A \cite{b79}, MGANs \cite{b69}, CNN \cite{b68}, GaitSet \cite{a2}, Gaitnet \cite{b58}, GaitPart \cite{b83} and Gaitref \cite{b85}, relying on visual appearance for recognition. The assessment includes Rank-1 Accuracy for normal walking (NM) at various camera viewpoints ($54^{\circ}$, $90^{\circ}$, and $126^{\circ}$), along with the mean accuracy across these viewpoints.}

\textcolor{black}{
Simultaneously, the evaluation covers model-based approaches, including PoseGait \cite{m5}, GaitGraph \cite{m8}, GaitGraph2 \cite{b84}, and our proposed BiGRU-dualStack model. Model-based methods aim to leverage inherent structures and dependencies within gait data for recognition. Table \ref{CASIA-B table} summarizes the Rank-1 Accuracy results for the considered models. Notably, in the appearance-based category, GaitSet \cite{a2}, GaitPart \cite{b83} and Gaitref \cite{b85} exhibit high accuracy, with the proposed BiGRU-dualStack model demonstrating competitive performance. In the model-based approach, our proposed Siamese biGRU-dualStack outperforms the other compared methods in terms of accuracy across different camera viewpoints.
} {In comparison with other state-of-the-art methods such as GaitMixer, notable differences emerge in terms of dataset scope, model architecture, and evaluation metrics. GaitMixer achieved a mean Rank-1 accuracy of 95.8\% (NM) on the CASIA-B dataset across angles ($54^{\circ}$, $90^{\circ}$, and $126^{\circ}$), considering 60 frames from the middle of the sequence data. In contrast, our BiGRU-dualStack method considered only 6 frames. Specifically, GaitMixer achieved a high Rank-1 accuracy on CASIA-B across $54^{\circ}$ and $90^{\circ}$, slightly outperforming our BiGRU-dualStack. However, our method demonstrated its robustness by excelling on a broader range of datasets, including SZU, OU-MVLP, and Gait3D.}

\begin{table}[ht]
\renewcommand{\arraystretch}{1.7}
  \begin{center}
    \caption{Rank-1 Accuracy of Different Models on CASIA-B Dataset.}
    \label{CASIA-B table}
    \begin{tabular}{|l|c|c|c|c|c|}
     \hline
   
       {Type}   & Model  & $54^{\circ}$ & $90^{\circ}$ & $126^{\circ}$ & Mean\\
      \hline 
       {} & VTM \cite{b77} & 55 &	46 & 54 & 51 \\
       
       {} & ViDP \cite{b78} &	64.2 &	60.4 & 65 & 63.2  \\
  
       {} & LRDF \cite{b70} &	77.7 &	59.9 &	75 & 70.9 \\
  
       {Appearance-} & C3A \cite{b79} & 75.7 & 63.7 &	74.8 & 71.4 \\

       {based} & MGANs \cite{b69} & 84.2 & 72.3 &	83 & 79.8  \\
       {} & CNN \cite{b68} & 94.6 & 88.3 & 93.8 &	92.2 \\
       
       {} & GaitSet \cite{a2} & 96.9 & 91.7 & 97.8 & 95.5 \\
       
       {} & Gaitnet \cite{b58} & 95.6 & 92.6 & 96 & 92.6 \\
       
       {} & GaitPart \cite{b83} &	98.5 & 92.3 & 98.4 & 96.4 \\
       
       {} & GaitRef \cite{b85} & 98.0 & 97.0 & 99.4 & 98.1 \\
       \hline
       
       {} & PoseGait \cite{m5} &	75 & 68.2 & 72.9 & 72\\
       {Model-} & GaitGraph \cite{m8} & 92.5 & 86.5 & 89.2 &	89.4\\
       {based} & GaitGraph2 \cite{b84}  & 85.6 & 81.5 & 83.2 &	83.4 \\
       {} & \textbf{BiGRU-dualStack} &	\textbf{95.6} &	\textbf{96.1} &	\textbf{95.5} &	\textbf{95.7}\\
      \hline
    \end{tabular}
  \end{center}
\end{table}

Additionally, Table \ref{SZU-table} shows the comparison of our Siamese BiGRU-dualStack approach on the SZU RGB-D Gait dataset with the GEI+PCA \cite{b80}, GES \cite{b71}, SPAE \cite{b72} and  \textcolor{black}{GaitNet \cite{b58}} methods, which are well-known approaches in gait recognition. The models were trained using the gait data of the first 49 subjects and the rest were used for testing. 

\begin{table}[ht]
  \centering
  \renewcommand{\arraystretch}{1.7}
  \caption{Comparisons of Different Models on SZU Dataset.}
  \label{SZU-table}
  \begin{tabular}{|c|c|}
    \hline
    Methods & Rank-1 Accuracy\\
    \hline
    GEI+PCA \cite{b80} & 27.0\\
    \hline
    GES \cite{b71} & 8.0\\
    \hline
    SPAE \cite{b72}  & 68.0 \\
    \hline
    GaitNet \cite{b58} & 85.2 \\
    \hline
    \textbf{BiGRU-dualStack} & \textbf{94.4} \\
    \hline
  \end{tabular}
\end{table}

According to the results in Table \ref{SZU-table}, our Siamese biGRU-dualStack achieved higher accuracy compared to the GEI+PCA, GES, SPAE, and GaitNet methods on the SZU RGB-D Gait dataset. This suggests that our proposed approach is effective in handling the RGB-D gait data and extracting discriminative features for accurate recognition.

\begin{table}[ht]
  \centering
  \renewcommand{\arraystretch}{1.7}
  \caption{Comparisons of Different Models on OU-MVLP Dataset.}
  \label{OU-MVLP}
  \begin{tabular}{|c|c|c|c|c|c|}
    \hline
    Methods &  $0^{\circ}$ & $30^{\circ}$ & $60^{\circ}$ & $90^{\circ}$ & Mean  \\
    \hline
    GEINET \cite{b86} & 8.2 & 32.3 & 33.6 & 28.5 & 25.7 \\
    \hline
    CNN-LB \cite{b68} & 14.2 & 32.7 & 32.3 & 34.6 & 28.5 \\ 
    \hline
    3in+2diff \cite{b87} & 25.5 & 50 & 45.3 & 40.6 & 40.4 \\
    \hline
    DigGAN \cite{b88} & 30.8 & 43.6 & 41.3 & 42.5 & 39.6 \\
    \hline
    GaitSet \cite{a2} & 77.7 & 86.9 & 85.3 & 83.5 & 83.4 \\
    \hline
    CNN-Pose \cite{b2} & 47.3 & 69.1 & 73.2 & 49 & 59.7 \\
    \hline
    SCN \cite{b90} & 78.6 & 87.4 & \textbf{85.9} & 83.2 & 83.8 \\
    \hline
   \textbf{BiGRU-dualStack} & \textbf{87.4} & \textbf{90.8} & 85.3 & \textbf{87.5} & \textbf{87.7} \\
    \hline
  \end{tabular}
\end{table}

\begin{table}[ht]
  \centering
  \renewcommand{\arraystretch}{1.7}
  \caption{Comparisons of Different Models on Gait3D Dataset.}
  \label{Gait3D-table}
  \begin{tabular}{|c|c|}
    \hline
    Methods & Rank-1 Accuracy\\
    \hline
    GaitSet \cite{a2} & 36.7\\
    \hline
    GaitPart \cite{b83} & 28.2\\
    \hline
    GaitGL \cite{b501} & 29.7 \\
    \hline
    GaitBase \cite{bd3} & 64.6 \\
    \hline
    \textbf{BiGRU-dualStack} & \textbf{86.6} \\
    \hline
  \end{tabular}
\end{table}

\textcolor{black}{To evaluate the proposed method's generalization, an experiment was conducted on the OU-MVLP dataset \cite{b721}, known as the largest public gait dataset. The dataset encompasses a substantial number of objects, and the comparison results are displayed in Table \ref{OU-MVLP}. Following strict adherence to the protocol under cross-view conditions, four typical views (0°, 30°, 60°, 90°) were utilized for the gallery set. Notably, the proposed BiGRU-dualStack method outperforms other models across different camera viewpoints, showcasing its robustness and effectiveness in gait recognition scenarios.}
\textcolor{black}{
Moreover, the BiGRU-dualStack method's superior performance is further highlighted in the Gait3D dataset. Table \ref{Gait3D-table} shows that the BiGRU-dualStack method achieved an impressive Rank-1 accuracy of 86.6\%, outperforming other models such as GaitSet \cite{a2}, GaitPart \cite{b83}, GaitGL \cite{b501}, and GaitBase \cite{bd3}. This result reinforces the method's robustness and effectiveness in gait recognition scenarios on the Gait3D dataset.}
\textcolor{black}{Overall, our empirical results underscore the Siamese BiGRU-dualStack as a promising and versatile model for gait recognition, capable of achieving state-of-the-art accuracy across diverse datasets and scenarios.}

\subsection{Ablation Study}
The study aimed to evaluate the impact of different recurrent neural network (RNN) architectures on the CASIA-B($90^{\circ}$) and SZU datasets. Specifically, we investigated the performance of RNN, LSTM, GRU, Bidirectional RNN \cite{b81}, Bidirectional LSTM \cite{b82}, Bidirectional GRU, 2-stacked Bidirectional RNN, 2-stacked Bidirectional LSTM and biGRU-dualStack. Table \ref{Ablation-table} presents the performance results of the ablation study for the CASIA-B and SZU datasets.

\begin{table}[ht]
\renewcommand{\arraystretch}{1.7}
  \begin{center}
    \caption{Results of Ablation Study for SZU and CASIA-B Dataset.}
    \label{Ablation-table}
    \begin{tabular}{|l|c|c|c|c|}
     \hline
      Methods   &  \multicolumn{2}{c|}{SZU} & \multicolumn{2}{c|}{CASIA-B ($90^{\circ}$)}\\
      \cline{2-5} 
      {}   & Accuracy  & Loss & Accuracy & Loss\\
      \hline 
       RNN & 81.57 & 14.33 & 75.31 & 18.96 \\
       \hline 
       LSTM & 92.53 & 11.13 & 90.33 & 13.31 \\
       \hline 
       GRU & 90.01 & 11.95 & 82.35 & 15.19 \\
       \hline 
       Bidirectional RNN & 84.06 & 14.07 & 85.42 & 12.89 \\
       \hline 
       Bidirectional LSTM & 92.60 & 11.27 & 88.54 & 14.87 \\
       \hline 
       Bidirectional GRU & 90.01 & 11.29 & 81.31 & 15.13 \\
       \hline 
       dualStack Bi-RNN & 91.09 & 9.30 & 87.07 & 12.05 \\
       \hline 
       dualStack Bi-LSTM & 94.28 & 9.34 & 96.07 & 9.14 \\
       \hline 
       \textbf{BiGRU-dualStack} & \textbf{94.44} & \textbf{7.51} & \textbf{96.08} & \textbf{7.88} \\
      \hline
    \end{tabular}
  \end{center}
\end{table}

The basic RNN architecture achieved lower accuracy compared to other architectures, indicating that it struggles to capture and utilize long-term dependencies in gait sequences. The vanishing gradient problem is a common issue with basic RNNs, which hampers their ability to effectively model long-term dependencies. On the other hand, LSTM (Long Short-Term Memory) networks outperformed the basic RNN architecture. LSTMs are designed to overcome the vanishing gradient problem by incorporating a memory cell and gating mechanisms, which enable them to retain and propagate relevant information across long sequences. The higher accuracy (SZU: $92.53\%$, CASIA-B($90^{\circ}$): $90.33\%$) and lower loss (SZU: $11.13\%$, CASIA-B($90^{\circ}$): $13.31\%$) suggest that LSTMs are effective at capturing the complex temporal dynamics in gait sequences. However, GRU (Gated Recurrent Unit) networks achieved competitive results, but slightly lower than LSTMs. GRUs have a simplified gating mechanism compared to LSTMs, resulting in fewer parameters. While they may not capture long-term dependencies as effectively as LSTMs, they can still model temporal dependencies reasonably well.

Bidirectional variants of RNN, LSTM, and GRU architectures incorporate information from both forward and backward directions of the input sequence. This allows them to leverage past and future context simultaneously, leading to a more comprehensive representation of gait patterns. Consequently, the bidirectional variants generally outperformed their unidirectional counterparts, achieving higher accuracy and lower loss. Furthermore, adding multiple stacked layers further enhances the model's capacity to learn complex representations. The dual-stack architectures, whether RNN, LSTM, or GRU, demonstrated improved performance compared to their single-layer counterparts. The additional layers enable the model to capture more intricate temporal dependencies and achieve better discriminative power for gait recognition.

Considering the performance metrics and the goal of accurate gait recognition, the biGRU-dualStack architecture was chosen as it achieved the highest accuracy with the lowest loss on the SZU and CASIA-B datasets. The bidirectional nature of GRU layers and the utilization of dual stacking contribute to its ability to effectively model gait patterns from both directions and capture complex temporal dependencies.

\section{Discussion}
\begin{figure*}[ht]
    \centering
    \includegraphics[width=12cm]{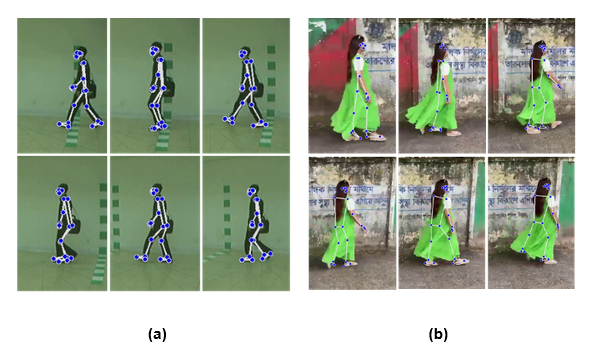}
    \caption{Landmarks on Human Subjects with Loose Clothing and Carrying Bags.}
    \label{fig7}
\end{figure*}
In this research paper, we investigate the application of the Siamese BiGRU-dualStack architecture for human gait recognition using gait landmarks. The primary goal of this study was to explore the performance of the proposed approach in various scenarios and assess its robustness and accuracy in challenging real-world conditions. The evaluation was conducted on a diverse dataset encompassing human subjects with different clothing types and individuals carrying bags or other external objects. Overall, our findings demonstrated the potential of the Siamese BiGRU-dualStack approach, coupled with contrastive loss, as a promising technique for accurate human pose estimation. The model showcased proficiency in detecting landmarks and inferring complex gait recognition under various conditions. However, there were specific challenges and limitations that surfaced during our experimentation. We address these challenges and delve into three essential aspects of our research.
\newline
\textbf{ \textit{Impact of Landmark Selection:}} We examine the influence of landmark selection on the accuracy and robustness of human gait identification. The human gait is a unique biometric characteristic that can be used for person recognition. We employed a state-of-the-art landmark detection system, such as Mediapipe, to extract anatomical landmarks from gait sequences.

Initially, we considered a wide range of landmarks, from 0 to 32, with $x, y, z$ coordinates, to assess their impact on the identification process. However, to investigate the efficiency and practicality of using fewer landmarks, we focused on two subsets: 11 to 32 (encompassing the upper body, from shoulder to foot) and 23 to 32 (solely the lower body). Our goal was to determine if utilizing a reduced number of landmarks would still yield reliable results, potentially simplifying the data acquisition process. 

{
Table \ref{probe} illustrates a slight decrease in accuracy for landmarks 11 to 32 and 23 to 32 across various probe scenarios on the CASIA-B dataset. However, our experiments revealed that the reduction in the number of landmarks did not significantly affect the overall accuracy of human gait identification. This indicates that even with fewer landmarks, reliable gait identification is achievable, which can simplify the data acquisition process without compromising accuracy significantly.} 

This suggests focusing on specific body regions for gait recognition may work well in real-world scenarios where capturing a full set of landmarks is difficult.
\begin{table}[ht]
\renewcommand{\arraystretch}{1.7}
  \begin{center}
    \caption{Rank-1 Accuracy on Different Ranges of Landmarks for CASIA-B Dataset.}
    \label{probe}
    \begin{tabular}{|l|c|c|c|}
     \hline
      {}   &  \multicolumn{3}{c|}{Landmarks}\\
      \cline{2-4} 
       {Probe}   & 0-32 & 11-32 & 23-32 \\
      \hline 
       $54^{\circ}$ & 95.6 & 94.7 & 93.8 \\
       \hline 
       $90^{\circ}$ & 96.1 & 95.5 & 95.2 \\
      \hline 
       $126^{\circ}$ & 95.5 & 95.2 & 95.2 \\
        \hline 
       Mean  & 95.7 & 95.1 & 94.7 \\
     \hline 
    \end{tabular}
  \end{center}
\end{table}

\textcolor{black}{When dealing with a mix of outdoor and indoor images, the complexity of contextual information can increase due to variations in lighting, background, and other environmental factors. However, by concentrating on personal landmarks, we can enhance accuracy. This approach focuses on unique and stable features of an individual's gait rather than fluctuating external conditions. As demonstrated in Table \ref{Gait3D-table} with the Gait3D dataset, this method can lead to improved performance in gait recognition.}

\textbf{ \textit{Impact of Clothing or Bags:}} We also explore the impact of clothing, specifically loose cloth, and objects like bags, on the accuracy and reliability of human gait identification using the Mediapipe landmark detection system.

During our experiments, we observed that landmark detection and subsequent gait identification performed exceptionally well on individuals wearing regular clothing. Human pose estimation algorithms, such as Mediapipe, heavily rely on detecting specific body landmarks to accurately infer body postures and movements. While the system has demonstrated impressive performance on human subjects wearing regular clothing, it is essential to assess its effectiveness when dealing with individuals wearing loose or baggy clothing. However, we encountered challenges when dealing with subjects wearing loose clothing.

{The landmark detection using MediaPipe for these individuals was slightly reduced compared to subjects in regular attire. The loose fabric of the clothing tended to obscure some key body landmarks, leading to reduced accuracy in certain poses. Consequently, the pose estimation algorithm exhibited challenges in accurately tracking body joints and postures in such instances. Similarly, carrying objects such as bags also impacted the accuracy and reliability of human gait identification using MediaPipe as it led to missing some landmarks. However, when certain body parts were partially obstructed by the bags, the algorithm efficiently inferred the missing landmarks based on their spatial relationships with other detected joints. Table \ref{NM BG CL} summarizes the accuracy of individuals in normal walking (NM), walking while carrying a bag (BG) and walking with a
coat (CL) at different angles. The mean accuracy drops from 95.7\% without a bag to 95\% with a bag and  95.2\% with a coat. This indicates a minor reduction in performance when bags are present.}

\begin{table}[ht]
\renewcommand{\arraystretch}{1.7}
  \begin{center}
    \caption{Rank-1 Accuracy on CASIA-B at Different Angles and Conditions.}
    \label{NM BG CL}
    \begin{tabular}{|l|c|c|c|c|}
     \hline
      {}   &  \multicolumn{4}{c|}{Probe}\\
      \cline{2-5} 
       {Conditions}   & $54^{\circ}$ & $90^{\circ}$  & $126^{\circ}$ & mean \\
      \hline 
       NM & 95.6 & 96.1   & 95.5 &  95.7\\
       \hline 
       BG  & 94.3 & 95.6  & 95.3  &  95\\
      \hline 
       CL  & 93.2 & 94.7 &  94.6 & 94.2 \\
      \hline 

    \end{tabular}
  \end{center}
\end{table}

To mitigate this issue, possible strategies can be explored, such as incorporating additional pre-processing steps to account for the presence of loose clothing or investigating alternative algorithms better suited for handling occluded landmarks. Furthermore, data augmentation techniques could be employed during training to simulate diverse clothing scenarios, thereby enhancing the algorithm's robustness to pose estimation under varying clothing conditions. Figure \ref{fig7} demonstrates the impact of using the Mediapipe pose estimation algorithm in two specific scenarios. Sub-figure (a) showcases human subjects carrying bags, with the algorithm accurately tracking body landmarks even in the presence of external objects. The colored dots represent the detected landmarks, providing insights into the algorithm's behavior under different conditions. Sub-figure (b) depicts human subjects wearing loose clothing, where the performance of the algorithm shows slight saturation due to occlusions caused by the loose fabric. 

\textbf{\textit{Understanding the Factors Contributing to Missed Identifications:}} During the pose estimation process, certain images may exhibit landmark saturation, where the extracted landmark values reach extreme or unusually high values. Such saturation can be caused by several factors, such as challenging lighting conditions, image quality, or the complexity of the pose. In such cases, the landmark values may no longer accurately represent the underlying body joint positions, leading to potential inaccuracies in the pose estimation. The saturation obscures the subtle differences between landmarks, making it challenging for the model to distinguish between various body parts accurately. Moreover, when the saturation is severe, normalization can compress the landmark values, reducing the variation between them. As a result, after normalization, certain body joint distinctions become indistinguishable from the model, leading to decreased accuracy of human identifications.

To address this issue, further research could focus on exploring advanced normalization techniques that adapt to the level of saturation in each image or investigating alternative loss functions that account for the saturation-induced variations in landmark values. Moreover, our findings highlight the need for continued research and development to optimize the performance of gait identification systems under different clothing conditions. This is particularly crucial for real-world applications, as individuals often wear various types of clothing in different environments. Enhancing the robustness of the system when dealing with such variations will ensure its effectiveness in practical scenarios, such as surveillance in crowded public spaces or law enforcement applications.

\section{Conclusion}
In this paper, we focused on gait recognition, aiming to achieve high accuracy while minimizing computational power and time requirements. However, stacked bidirectional LSTM and stacked bidirectional GRU architectures exhibited better performance than the basic RNN, with GRU slightly trailing behind LSTM. Hence, we proposed the Siamese biGRU-dualStack approach, which outperformed state-of-the-art methods on the CASIA-B, SZU RGB-D, OUMVLP, and \textcolor{black}{Gait3D} datasets. Our model effectively captured important gait features, resulting in superior recognition performance. Additionally, we incorporated the use of MediaPipe landmark collection, which further enhanced the model's ability to capture complex gait patterns. To mitigate the impact of variations in angles of view and body orientations, we employed procrustus analysis, which allowed for a more accurate comparison, and gait analysis. Overall, the Siamese biGRU-dualStack approach shows promise for practical applications in biometric identification and surveillance systems, providing accurate gait recognition with reduced computational requirements.


\end{document}